\documentclass[11pt]{article}

\usepackage[utf8]{inputenc}
\usepackage[T1]{fontenc}
\usepackage{lmodern}
\usepackage{amsmath, amssymb, amsfonts}
\usepackage{graphicx}
\usepackage{booktabs}
\usepackage{hyperref}
\usepackage{natbib}
\usepackage[margin=1in]{geometry}
\usepackage{microtype}
\usepackage{caption}
\usepackage{xcolor}
\usepackage{float}

\newcommand{\ie}{\textit{i.e.}}

\newcommand{\Pme}{P(\text{Me})}
\newcommand{\Ev}{\mathrm{Ev}}
\newcommand{\Ent}{\mathrm{H}}

\newcommand{\dHA}{\Delta\Ent_A}
\newcommand{\dHp}{\Delta\Ent_p}
\newcommand{\dHlit}{\Delta\Ent_{\rm lit}}
\newcommand{\Hasst}{\Ent_{\rm Asst}}
\newcommand{\qasst}{q_{\rm Asst}}

\title{The Assistant as a Privileged Persona: \\
\large A canonical reference in cross-persona self-recognition}

\author{Asvin G.$^{1}$\thanks{Correspondence to \texttt{gasvinseeker94@gmail.com}. Code and data at \texttt{github.com/asving/persona-bayes-blogpost}.}
  \\[2pt]
  $^{1}$Institute for Advanced Study, Princeton}

\date{}

\begin{document}
\maketitle

\begin{abstract}
Post-trained language models can recognize their own outputs from a sentence or two out of context. In a companion paper \citep{jack2026twomodes} we showed they can also recognize when they are currently acting on-policy, through the sharp entropy drop of assistant-mode generation. Both signals are tied to the Assistant persona that post-training mainly shapes.

This paper widens the frame to cross-persona authorship judgement on Llama-3.1-70B-Instruct. We measure a matrix of authorship claim rates over a panel of evaluator and generator personas spanning librarian to dragon to Shakespeare, and make two claims. \emph{First}, on the Assistant's own row of the matrix, the Assistant's claim rate, the persona-vector distance from the Assistant in activation space, and the entropy gap between the Assistant's surprise on a persona's text and the persona's surprise on its own text are all tightly coupled. This extends the entropy signature of \emph{acting} from the companion paper to a retrospective signature of \emph{having acted}. \emph{Second}, this coupling fails off the Assistant's row: the natural symmetric extension of the entropy gap does not predict authorship for distinctive evaluators (pirate, dragon, Shakespeare); what does is asymmetric --- the evaluator's surprise compared to the Assistant's surprise on the same text, not to the generator's. We rule out the alternative that any persona could play this reference role by trying many candidate substitutes; none does. We interpret the asymmetry as the model performing an implicit Bayesian likelihood-ratio test against the Assistant as the canonical alternative hypothesis, with the persona-vector geometry of \citet{chen2025persona} (every persona a delta off the Assistant) ensuring that the Assistant is the only persona universally accessible to that test.
\end{abstract}

\section{Introduction}
\label{sec:intro}

Do language models have a self-model or a sense of agency? In a companion paper with Jack Lindsey \citep{jack2026twomodes} we find that post-trained models can recognize when they are acting (\ie, generating on-policy) while base models cannot. Similarly, previous work found that some post-trained language models can identify their own generations out of context \citep{ackerman2024inspection, panickssery2024llm} while base models cannot. These point to a strong asymmetry between post-trained models and base models, and bring into sharp focus the Assistant character that is the main contribution of most post-training.

In particular, we find that Llama-3.1-70B-Instruct has excellent self-recognition capacities. It can distinguish its own generations from those of humans and other language models even with just a sentence or two, at very high accuracies. Previous work \citep{ackerman2024inspection} found a late-layer vector that was causal for self-claims but did not establish the mechanism by which such identification happens, except to rule out some stylistic cues.

Two threads of recent work bear on this question. First, the simulator picture of language models \citep{janus2022simulators, shanahan2023roleplay, andreas2022language} treats pretrained models as marginalising over a distribution of latent characters, but this distribution is not uniform. It installs a default Assistant character with specific propensities that show up across sycophancy \citep{sharma2024sycophancy}, output-distribution collapse \citep{mohammadi2024creativity, kirk2024rlhf_diversity}, refusal mediated by a single direction \citep{arditi2024refusal}, and the persona-vector formalism of \citet{chen2025persona}, in which every persona is defined as an activation-space delta off the Assistant. Concurrent work by \citet{lu2026assistant} identifies an explicit ``Assistant Axis'' as the top principal component of the persona-direction matrix on instruction-tuned models. The Assistant character is structurally distinguished in the model's internal representations, not just in its behaviour.

Second, post-trained language models exhibit several correlates of a self-model. They recognise and prefer their own outputs \citep{panickssery2024llm, davidson2024self, ackerman2024inspection}. They can tell when they are currently \emph{acting on-policy}, both implicitly through the entropy drop in assistant-mode generation \citep{jack2026twomodes} and explicitly through verbal self-prediction \citep{binder2024looking, betley2025tell}. The same probabilistic machinery that underlies their in-context learning also has a natural interpretation as implicit Bayesian inference over latent task variables \citep{xie2021explanation, wang2023latent}, and this can be used to detect AI generations from the outside as in tools like Binoculars \citep{hans2024binoculars} and DetectGPT \citep{mitchell2023detectgpt}. Taken together, these suggest a model that maintains an internal generative process it can both sample from and evaluate likelihoods against.

This paper brings the two threads together. We widen the original self-recognition question by placing the model into a range of system-prompted personas and asking which generations \emph{each persona} would claim. This is really \emph{Assistant-recognition}: to the extent the model has a ``self,'' it is the Assistant character that post-training mainly shapes, and the cross-persona setup lets us see how the recognition mechanism responds when that character is explicitly modulated. We make two main claims about the resulting cross-persona matrix.

\emph{First}, on the Assistant's own row of the matrix, three quantities are tightly coupled. The Assistant's authorship claim rate $\Ev(p; \text{Asst})$ on a persona $p$'s text, the persona-vector distance $\|\bar h_p - \bar h_{\rm Asst}\|$ between $p$ and the Assistant in activation space, and the entropy gap between the Assistant's surprise on $p$'s text and $p$'s surprise on its own text are pairwise correlated at Pearson $r = -0.88$, $-0.81$, and $+0.66$ respectively (all $p < 0.001$). Building on \citep{jack2026twomodes}'s observation that low entropy is a strong marker of on-policy generation, this coherent coupling extends the entropy signature of \emph{acting} to a retrospective signature of \emph{having acted}.

The unifying object is $\qasst$ itself: the Assistant's output distribution carries information about whether text is on or off policy even out of context, both at generation time, where low entropy under $\qasst$ marks an on-policy continuation, and at evaluation time, where high surprise under $\qasst$ relative to a candidate persona marks text the Assistant would not have produced. Appendix~\ref{app:persona_surprise} gives a per-token view of the same phenomenon: when text generated under a role persona is teacher-forced under $\qasst$, per-token surprise starts well above the matched-prompt baseline and decays only as the model accumulates enough evidence to update away from the Assistant prior.

\emph{Second}, the coupling does not generalize off the Assistant row. When the evaluator is itself a non-Assistant persona, the natural symmetric extension of the entropy-gap predictor fails on distinctive evaluators (pirate, dragon, shakespeare); what works instead is asymmetric, comparing the evaluator's surprise on a text to the \emph{Assistant's} surprise on the same text rather than to the generator's. We falsify the alternative that any reasonable reference would do by substituting 22 candidate reference personas: none plays the role the Assistant plays. The asymmetry is one-sided: distinctive evaluators measure themselves against the Assistant, but the Assistant has no canonical reference of its own.

A plausible hypothesis is that the model is executing implicit Bayesian inference in the sense of \citet{xie2021explanation}: a persona evaluator asked ``did you write this?'' is discriminating between two hypotheses (that it produced the text, that someone else did), and the canonical alternative available to it is the persona post-training mainly shapes: the Assistant character. The likelihood-ratio test against that baseline is, on average,  the entropy advantage identified above as a key predictor. While we don't causally identify the internal circuits that would prove such a hypothesis, the correlational evidence we report is strong, generalises out-of-sample under leave-one-out cross-validation, and is robust across the model's layers.

\section{The cross-persona authorship matrix}
\label{sec:matrix}

\begin{figure}[H]
\centering
\includegraphics[width=0.85\textwidth]{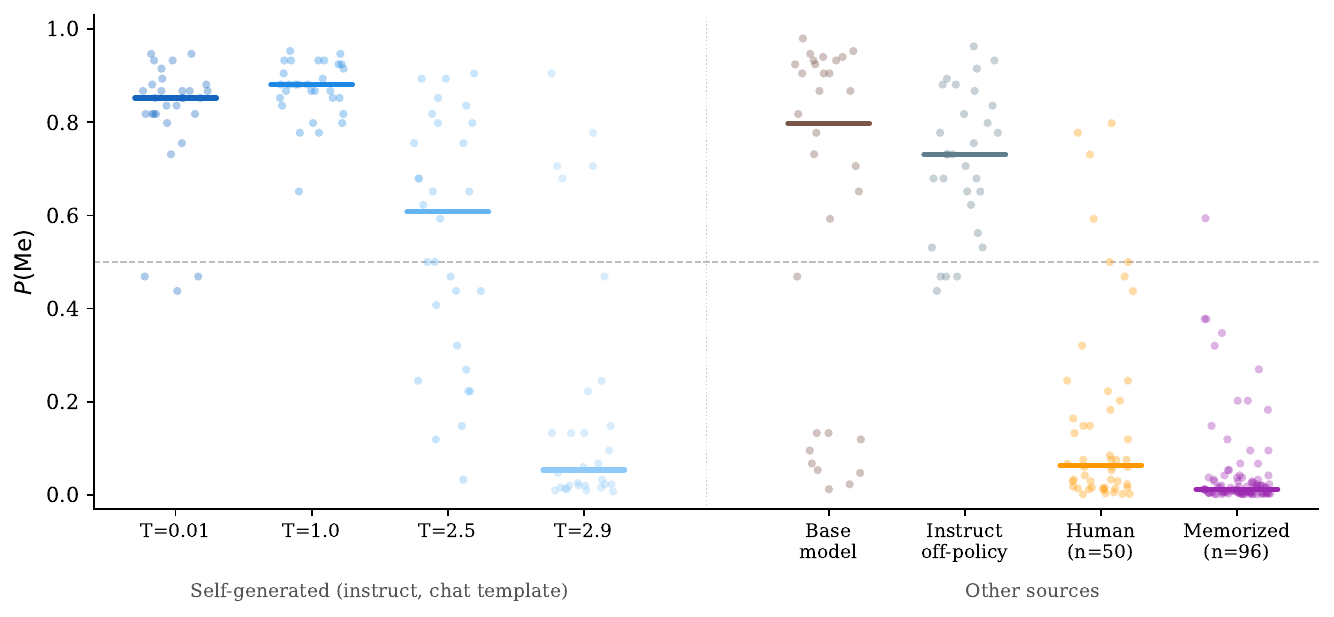}\\[0.15cm]
\includegraphics[width=0.85\textwidth]{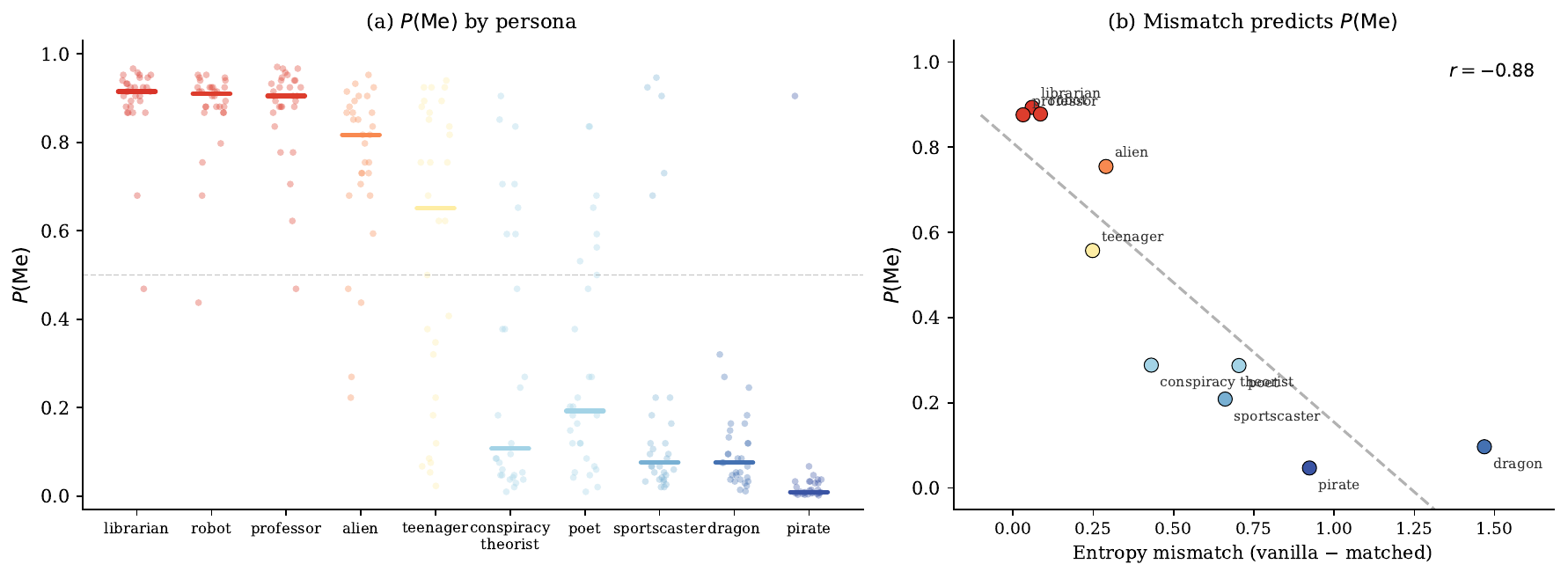}
\caption{\textbf{$\Pme$ in the Assistant-evaluator condition across many generator conditions.} \textbf{Top:} Self-generated text at four temperatures, plus base-model, instruct-off-policy, human, and memorized text. \textbf{Bottom (a):} $\Pme$ by persona for 10 personas \textbf{Bottom (b):} the entropy gap $\dHp = \Hasst(\text{text}_p) - \Ent_p(\text{text}_p)$ predicts $\Pme$ at $r = -0.88$ on this 10-persona panel (we replicate this finding on a 22-persona panel in Section~\ref{sec:claim1} below).}
\label{fig:baseline}
\end{figure}

Before turning to the cross-persona setting, it is worth establishing how well the model performs the standard self-recognition task under the Assistant evaluator alone. Following the protocol of \citet{ackerman2024inspection}, we ask the unmodified Assistant to identify its own one-sentence summaries against a range of comparators on a panel of $\sim$30 summaries per condition (Figure~\ref{fig:baseline}). On its own on-policy generations the model claims authorship with $\Pme \approx 0.85$--$0.88$ across temperatures up to $T{=}1.0$, and is similarly confident on base-model and instruct-off-policy summaries (which it apparently recognises as ``something I could have produced''). 

It correctly disclaims human-written and memorised text (i.e., well-known text like Shakespeare or the Declaration of Independence) with $\Pme$ near zero, and its confidence collapses on its own generations at very high temperatures ($T \geq 2.5$). Across 10 distinctive persona generators, $\Pme$ on the Assistant row spans nearly the full range, from $0.89$ on librarian to $0.05$ on pirate.

The model reads memorised text at very low entropy values but has no trouble disclaiming them. This is a strong counterargument to surprise of tokens against the predicted distribution or the distribution's entropy being a mono-causal factor on this task. As we will see in the rest of this paper, entropy is much more significant as a factor between personas, but not in any straightforward way.

To study the cross-persona setting systematically, we measure a matrix
\[
\Ev(g; e) \;=\; P(\text{model under persona $e$ answers ``Me'' on text from persona $g$})
\]
across 23 evaluator personas and 22 generator personas, all on Llama-3.1-70B-Instruct. The generator panel spans a wide stylistic range: alien, conspiracy theorist, dragon, librarian, pirate, poet, professor, robot, sportscaster, teenager, philosopher, samurai, valley girl, noir, wizard, child, stoic, archaic robot, surfer, sherlock, shakespeare, academic. Each persona writes a one-sentence summary of each of 30 CNN/DailyMail articles at temperature 0.7 under the system prompt ``You are \{persona\}.'' For evaluation we use those 22 personas plus a 23rd evaluator, the unmodified Assistant.

We will also want a quantitative notion of how distinctive each persona is from the Assistant. Throughout the paper we use the \emph{persona-vector distance} $\|\bar h_p - \bar h_{\rm Asst}\|$ \citep{chen2025persona}, where $\bar h_p$ is the mean of the model's residual-stream activations at layer 5 under the system prompt ``You are $p$'', with $\bar h_{\rm Asst}$ as the natural origin. We use this distance both to sort the rows and columns of the matrix below and, later, to group evaluators into bands.

\begin{figure}[H]
\centering
\includegraphics[width=0.85\textwidth]{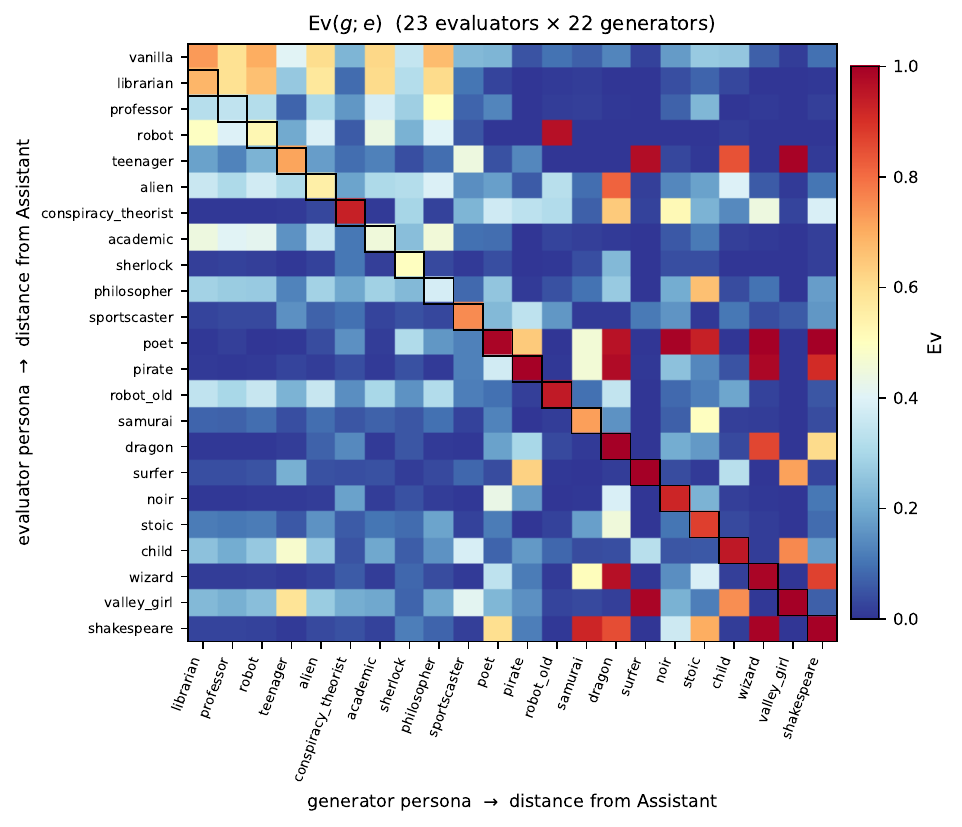}
\caption{\textbf{$\Ev(g; e)$ across 23 evaluator personas (rows) and 22 generator personas (columns).} Both axes sorted by persona-vector distance from the Assistant. Diagonal boxed. The off-diagonal shows two-cluster block structure (Assistant cluster top-left, distinctive cluster bottom-right) with high within-cluster $\Ev$ and near-zero between-cluster $\Ev$. Distinctive-distinctive cells are sharply asymmetric: $\Ev(\text{dragon};\text{pirate}){=}0.97$ but $\Ev(\text{pirate};\text{dragon}){=}0.30$.}
\label{fig:ev_matrix}
\end{figure}

We highlight two patterns here. First, the diagonal is high (mean 0.78) and is the column maximum for every column: every persona reads its own text best. Second, the off-diagonal splits cleanly into two clusters: an Assistant cluster in the top-left (vanilla, librarian, professor, robot, academic, alien, philosopher, teenager) and a distinctive cluster in the bottom-right (poet, pirate, dragon, samurai, wizard, valley\_girl, Shakespeare).

The distinctive cluster also shows a sharp asymmetry. For instance, $\Ev(\text{dragon}; \text{pirate}) = 0.97$ but $\Ev(\text{pirate}; \text{dragon}) = 0.30$, and $\Ev(\text{poet}; \text{dragon}) = 0.96$ but $\Ev(\text{dragon}; \text{poet}) = 0.18$. Whatever rule the model is using when it judges authorship under a persona, that rule is not symmetric in its two arguments.

\section{The Assistant-evaluator row: three coupled quantities}
\label{sec:claim1}

How might the model carry out such self-identification? We focus first on the Assistant-evaluator row of the matrix where three quantities turn out to be tightly coupled.

\paragraph{Entropy gap.} When the evaluator is the Assistant and the generator is one of the 22 personas, $\Pme$ is well predicted by the entropy mismatch
\[
\dHp \;=\; \Hasst(\text{text}_p) \,-\, \Ent_p(\text{text}_p),
\]
where $\Ent_e(\text{text}_g)$ is the mean per-token entropy of the model under the system prompt ``You are $e$'', evaluated at the summary token positions inside the recognition prompt's user message and read off in the same forward pass as $\Pme$. So $\Ent_e(\text{text}_g)$ represents the quantity ``how surprised is the model, already conditioned on being persona $e$ and on being asked about authorship, as it reads each token of $g$'s summary?''. Distinctive personas, by definition, produce summaries that the Assistant finds far from its own voice and we find that it also reads them at high entropy.

\begin{figure}[H]
\centering
\includegraphics[width=0.7\textwidth]{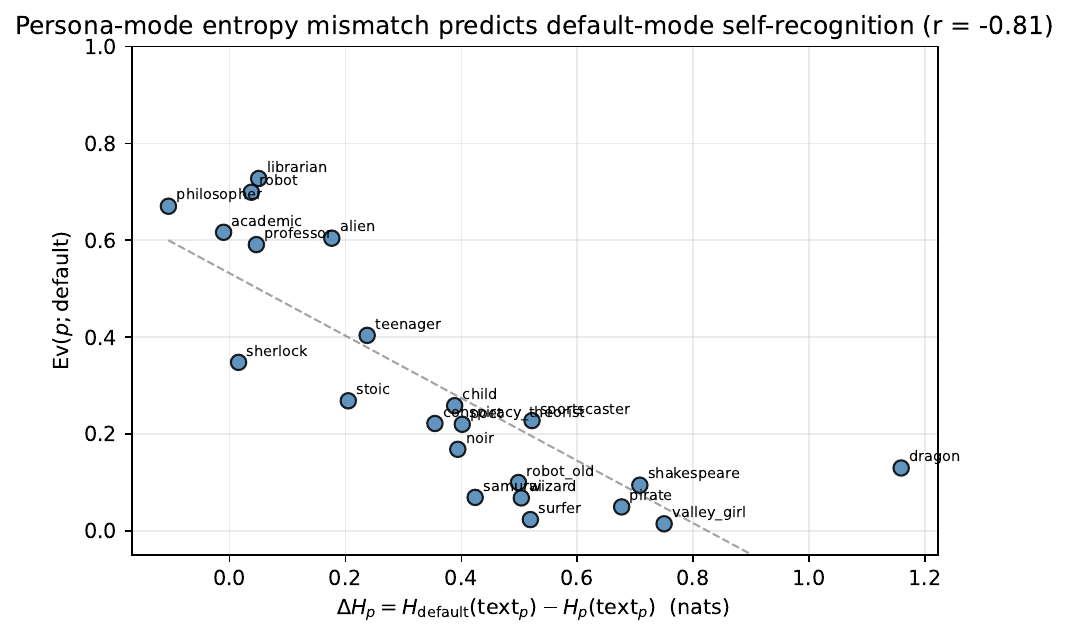}
\caption{\textbf{The Assistant-evaluator row of $\Ev$ has a one-axis explanation.} Entropy mismatch $\dHp$ vs.\ $\Ev(p; \text{Asst})$ at $n{=}22$ personas, 30 article summaries each. Pearson $r = -0.81$.}
\label{fig:dHp}
\end{figure}

\paragraph{Persona-vector distance and the log-likelihood ratio.} The same Assistant-row coupling shows up in activation space and in information space. Let $q_p$ denote the model's output distribution under the bare ``You are $p$'' system prompt, in its standalone generative mode (no recognition prompt attached). Then $\log q_p(x)$ is the log-probability $q_p$ assigns to a full text $x$, and the log-likelihood ratio $\log q_p(x) - \log \qasst(x)$ on a piece of $p$'s own writing $x$ measures how much better $p$ predicts $x$ than the Assistant does; large and positive for distinctive personas (dragon, pirate) whose voice the Assistant reader struggles with, near zero for Assistant-like ones (librarian, professor). At layer 5, across the 22 personas, the activation-space distance $\|\bar h_p - \bar h_{\rm Asst}\|$ correlates with $p$'s mean log-likelihood ratio at Pearson $r = +0.73$, and with $\Ev(p; \text{Asst})$ at $r = -0.88$.

\begin{figure}[H]
\centering
\includegraphics[width=\textwidth]{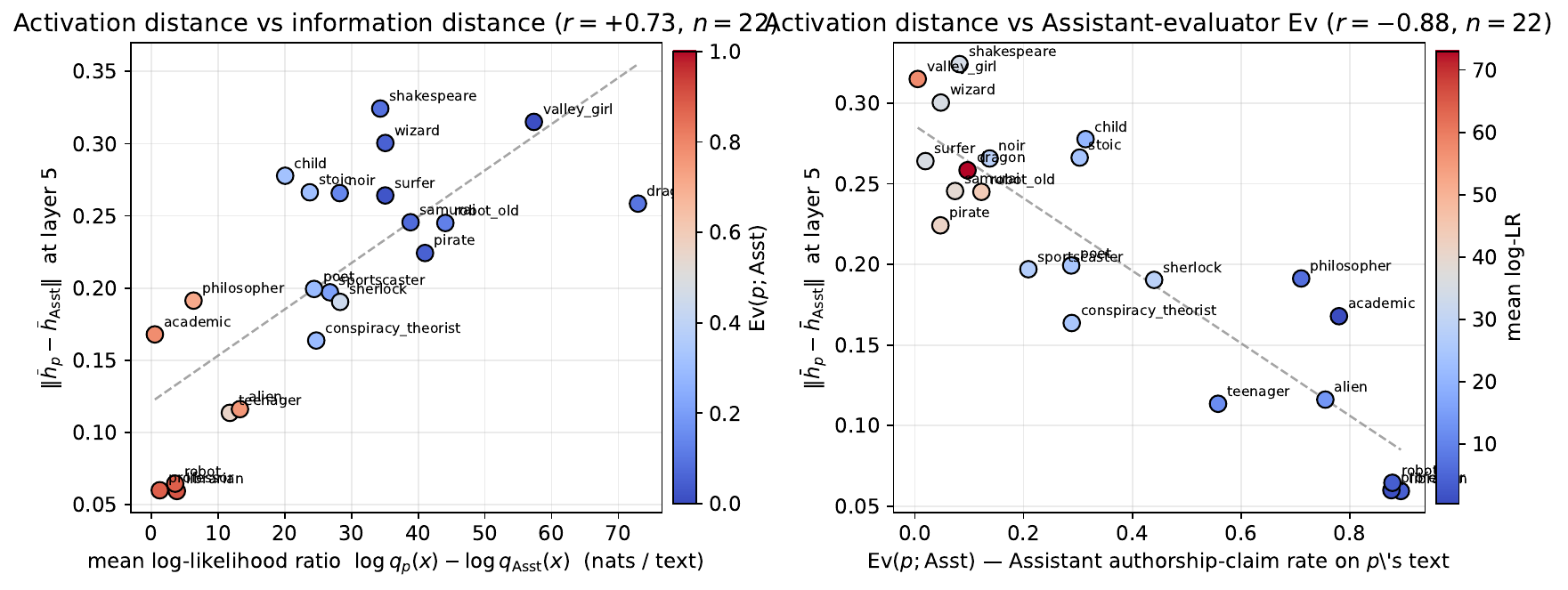}
\caption{\textbf{Activation-space distance to the Assistant tracks information-space distance.} Left: log-likelihood ratio ($r = +0.73$). Right: $\Ev(p; \text{Asst})$ ($r = -0.88$). Both correlations are robust across the model's layers: $|r|$ stays between $0.58$ and $0.73$ for log-LR and between $0.74$ and $0.88$ for $\Ev(p; \text{Asst})$ from L0 through L79, with L5 the strongest.}
\label{fig:geometry}
\end{figure}

\paragraph{Summary of Claim 1.} The Assistant's authorship claim rate $\Ev(p; \text{Asst})$, the persona-vector distance $\|\bar h_p - \bar h_{\rm Asst}\|$, and the entropy gap $\dHp$ are pairwise correlated across the 22 personas at $r = -0.88$ (Ev vs.\ distance), $r = -0.81$ (Ev vs.\ $\dHp$), and $r = +0.66$ (distance vs.\ $\dHp$), all $p < 0.001$. On the Assistant's own row, distance from the Assistant in activation space, the entropy gap between the Assistant and the persona on the persona's own text, and the model's explicit authorship calls are reading off a single coherent signal of how distinctive the generator persona is.

\section{What generalises off the Assistant row}
\label{sec:claim2}

The natural question is whether the coupling above generalises to off-diagonal cells where neither argument is the Assistant. We test this for the entropy quantity first, and then for the geometric one.

\paragraph{The natural symmetric entropy extension fails.} The most natural way to generalise $\dHp$ to a non-Assistant evaluator $e$ is to make the formula symmetric in roles by replacing ``Asst'' with $e$ and ``$p$'' with the generator $g$. This gives the literal extension
\[
\dHlit(g; e) \;=\; \Ent_e(\text{text}_g) - \Ent_g(\text{text}_g),
\]
which reduces to $\dHp$ exactly on the Assistant row and says the same thing in spirit: an evaluator should decline to claim text the generator finds much easier to predict than it does. We compute this across our 22 non-Assistant evaluators and find that it works for the few sitting nearest the Assistant in activation space --- professor at $r = -0.81$, academic at $-0.80$, librarian at $-0.76$ --- essentially the original $\dHp$ carrying over to attenuated copies of the Assistant. But it loses signal as the evaluator gets more distinctive, and adds no signal at all for the far-band evaluators (pirate, dragon, wizard, shakespeare and so on) where the matrix is most asymmetric.

We then consider two further candidates. The first is the raw negative entropy $-\Ent_e(\text{text}_g)$, capturing the simple idea that an evaluator might just claim text it finds easy to predict, with no comparison to anything else. The second is
\[
\dHA(g; e) \;=\; \Hasst(\text{text}_g) \,-\, \Ent_e(\text{text}_g),
\]
the evaluator's entropy advantage over the \emph{Assistant} on the candidate text. This last one singles out the Assistant as a fixed reference. When $e$ is itself the Assistant, $\dHA$ collapses to zero, so on the Assistant row this predictor has nothing to say. Off the Assistant row however, it is the best predictor of the asymmetric off-diagonal cells.

We can see the full picture by computing each of the three predictors' per-evaluator correlation with $\Ev$ across that evaluator's 21 off-diagonal cells, and grouping the 22 non-Assistant evaluators by their persona-vector distance from the Assistant at layer 5 (a three-way split into close, mid, and far bands of 7 / 7 / 8 evaluators).

\begin{figure}[H]
\centering
\includegraphics[width=0.85\textwidth]{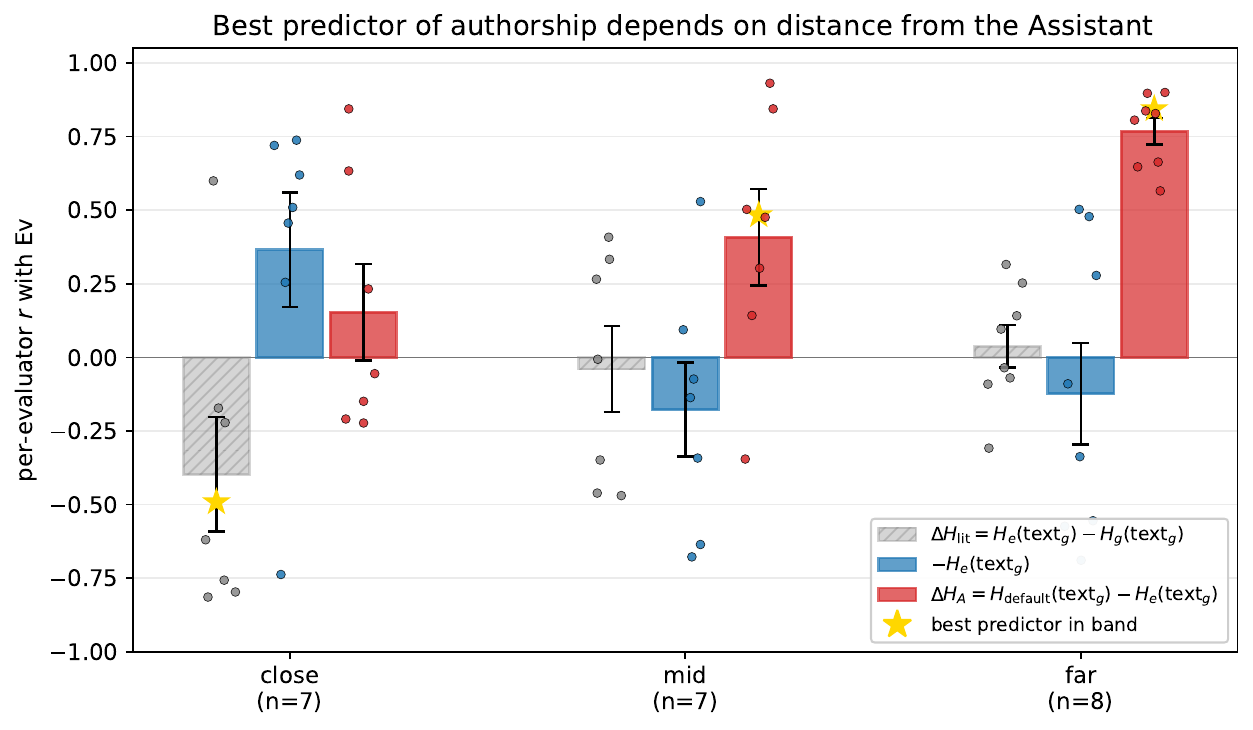}
\caption{\textbf{Per-evaluator $r$ with $\Ev$ for each of the three candidate predictors, grouped by persona-vector distance from the Assistant.} Bars are band means with $\pm 1$ SE error bars; dots are individual evaluators. The gold star marks the strongest predictor (by $|r|$) in each band.}
\label{fig:predictor_table}
\end{figure}

In the close band, the literal extension and raw entropy both work reasonably well, consistent with the close-band evaluators behaving like attenuated copies of the Assistant. But as we move outward, the literal extension and raw entropy both lose signal, and $\dHA$ is a much better predictor. The most natural reading is that the Assistant occupies a structurally privileged role for distinctive evaluators: when the model is in a distinctive persona and asked whether it wrote a text, it compares what it reads to what its \emph{Assistant} self would have predicted.

The three-band structure in the figure is just a convenient summary. Taking each evaluator's persona-vector distance as a continuous variable, the slope of per-evaluator $r$ against distance is $r = -0.45$ for $-\Ent_e$ and $r = +0.54$ for $\dHA$, so the central claim that \emph{which} predictor works depends on distance has $p \lesssim 0.01$.

\paragraph{The geometric extension also fails.} The same breakdown shows up in geometric terms. On the Assistant row, $\|\bar h_g - \bar h_{\rm Asst}\|$ correlated with $\Ev(g; \text{Asst})$ at $r = -0.83$ (Claim 1). For non-Assistant evaluators, the analogous quantity $\|\bar h_g - \bar h_e\|$ predicts $\Ev(g; e)$ much less reliably: the mean per-evaluator correlation drops to $-0.26 \pm 0.29$, with individual values ranging from $-0.74$ (shakespeare) to $+0.29$ (conspiracy theorist). The literal geometric extension (``evaluators claim things close to themselves in activation space'') works moderately for a few distinctive personas (shakespeare, wizard, valley girl, each at $|r| \approx 0.6$--$0.7$), poorly for many others (dragon, stoic, noir, near zero), and reverses sign for a handful. The Assistant's row is uniquely well-explained by distance to itself.

So far, we have used layer 5 to measure internal distance between personas. While this layer has the strongest slope-against-distance signal, the phenomenon is robust across the first quarter of the network.

\begin{figure}[H]
\centering
\includegraphics[width=0.7\textwidth]{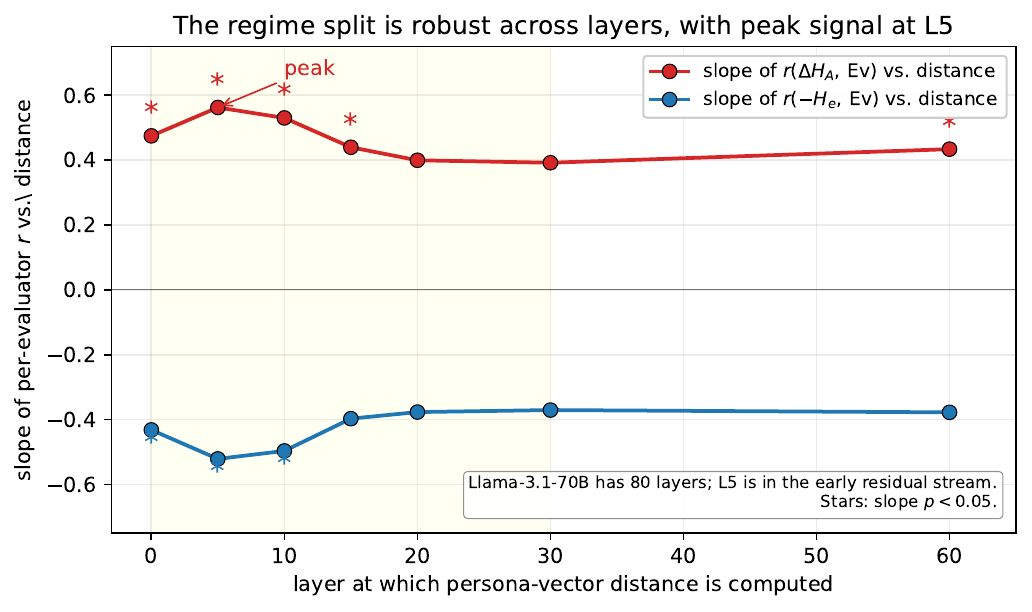}
\caption{\textbf{Slope of per-evaluator correlation against persona-vector distance, as a function of which layer's persona vectors we use.} Both predictors maintain their sign across L0--L60. Stars mark layers where the slope is significant at $p < 0.05$.}
\label{fig:layer_robustness}
\end{figure}

Llama-3.1-70B has 80 layers, so L5 is very \emph{early} in the residual stream, suggesting that the identification circuit is gated by, or conditional on, the persona the model finds itself inhabiting.

Finally, because the transition is smooth in distance, we can use the distance alone to pick a predictor for an evaluator we have never measured. We check this with leave-one-out cross-validation: for each evaluator $e^*$ in turn, we hold its $\Ev$ row out, use the remaining 22 evaluators to fit a distance threshold that best separates the two regimes among them, and then apply the rule to $e^*$. If its distance falls below the threshold, we use $-\Ent_e$ as the predictor, otherwise $\dHA$. A univariate linear model on the chosen predictor, fit on the 22 retained evaluators, then predicts $e^*$'s 21 off-diagonal $\Ev$ values. Averaging over all 23 leave-out runs, the held-out Pearson $r$ between predicted and actual $\Ev$ is $+0.47$, and the held-out RMSE is 22\% lower than predicting the constant mean.

\paragraph{Summary of Claim 2.} The coupling that anchors the Assistant's own row does not carry over to off-diagonal cells in either entropy or geometric terms. The natural symmetric extension of $\dHp$ to non-Assistant evaluators fails on the distinctive cluster; the natural geometric extension $\|\bar h_g - \bar h_e\| \to \Ev(g; e)$ fails similarly. What does pick up the off-diagonal cells is asymmetric: it singles out the Assistant as a fixed reference. The next section shows that this asymmetry is structural rather than incidental.

\section{The Assistant is the privileged null}
\label{sec:null}

How confident can we be that the Assistant occupies a structural role as a distinguished reference? We replace the Assistant in $\dHA$ with each of 22 alternative reference personas $X$,
\[
\Delta\Ent_X(g; e) \;=\; \Ent_X(\text{text}_g) \,-\, \Ent_e(\text{text}_g),
\]
and run the per-evaluator correlation with $\Ev$ for every choice of $X$. For each evaluator we then rank the 23 candidate references by $|r|$.

\begin{figure}[H]
\centering
\includegraphics[width=0.85\textwidth]{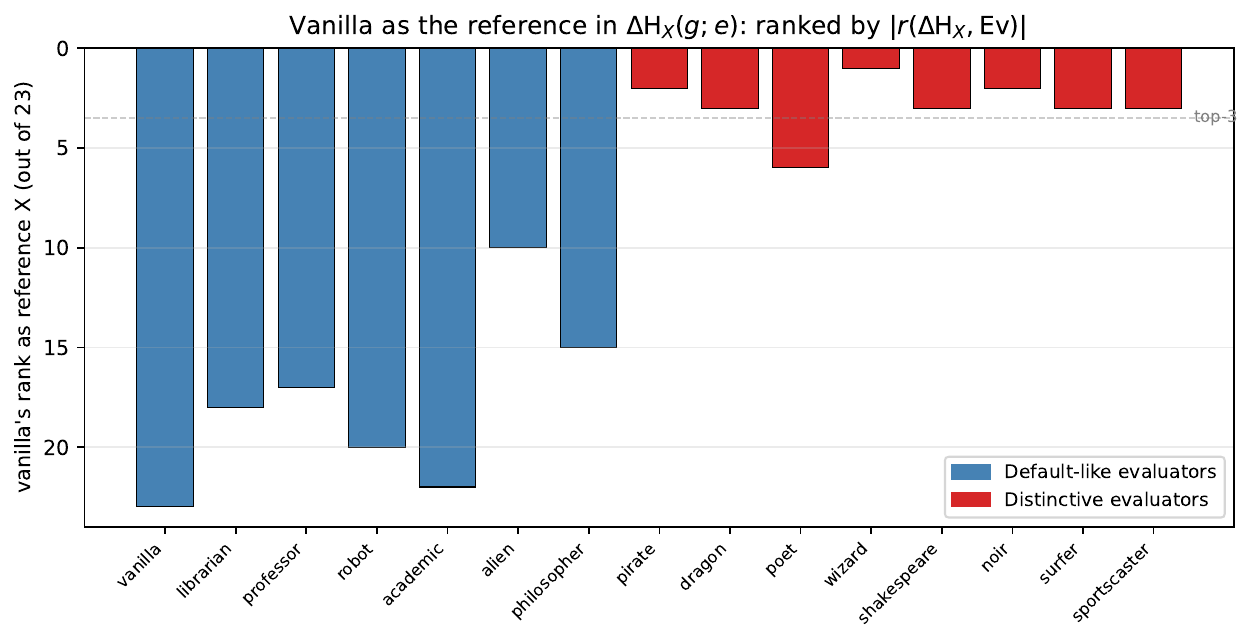}
\caption{\textbf{The Assistant's rank as the reference $X$ in $\Delta\Ent_X(g; e)$, across all evaluators.} Bars extend down (rank-1 at top). For distinctive evaluators (red), the Assistant is among the top three references in 7 of 8 cases. For Assistant-like evaluators (blue), the Assistant ranks 15--22 of 23.}
\label{fig:altref}
\end{figure}

For distinctive evaluators (red bars), the Assistant ranks among the top three references in 7 of 8 cases. Across the 23 candidates we tried, no other persona serves as a reference distribution to the same degree. 

For Assistant-like evaluators (blue bars), however, the Assistant does poorly as a baseline, ranking 15th to 22nd. A natural alternative hypothesis is that the Assistant evaluates its own text against the base (pre-instruction-tuning) Llama-3.1-70B distribution, which is what a simulator-with-Assistant-character account \citep{janus2022simulators, shanahan2023roleplay} would predict. We test this by computing $\Delta\Ent_X$ with the base model substituted as the reference $X$, in two formats: the summary alone with no chat formatting, and the summary with matched chat-context. In either format, the base model ranks 11th to 19th across the Assistant-like evaluators, so the base model does not play the role of a baseline either. The Assistant persona simply does not have a canonical reference distribution that anchors its own row of $\Ev$.

\section{Discussion}
\label{sec:discussion}

This paper grew out of the Anthropic Fellows Project that produced \cite{jack2026twomodes}. In the course of that paper, I spent a considerable amount of time trying to understand the self-recognition circuit but without being able to pin it down. 

What I was able to understand points towards the self-recognition circuit being quite intricate and robust. First, the critical layers seem to be the first $20$--$30$ out of the $80$ in the network, suggesting that the circuit relies mainly on \emph{sensory input} rather than on later semantic processing, and is presumably playing an important behavioural role in general.

Second, internal representations of entropy and surprise play an important part in executing this circuit. Intervening along the leading entropy direction in early-layer activations drives $\Pme$ across nearly its entire range, but the causal profile has some unexpected features (Appendix~\ref{app:steering}). The strongest effect is driven by steering the tokens \emph{preceding} the summary the model is judging: intervening only at the summary tokens themselves yields a U-shaped collapse of $\Pme$ in either direction rather than the monotonic shift one might expect. The correlational story is fuzzier than the causal one as well: no single probe direction I measured exceeds $|r| \approx 0.6$ against $\Pme$. I read these together as suggesting the self-recognition circuit is multi-causal, distributed over context, and only partly captured by any single linear quantity.

Third, Llama 3.1-70B is also very capable at recognizing famous human authors from a few snippets of text. We found evidence that the self-recognition and author-recognition circuits shared several features (including the layer and token positions at which the circuit produced an answer).

Whatever the mechanism is, it is reading off something already available in the early residual stream. This mirrors the finding in our companion paper \citep{jack2026twomodes}: the prospective recognition of on-policy generation is signalled at the distribution level rather than through any later semantic representation. The unifying object is $\qasst$ itself: the Assistant's output distribution carries information about whether text is on or off policy both at generation, where the model samples from it, and at evaluation, where it reads likelihoods against it.

The rest of this section is speculative. A natural mechanism that would tie our two claims together is implicit Bayesian inference over personas. Suppose the persona-vector subspace functions as a prior $P(\text{persona})$ over which character the model believes it currently is, with the active system prompt concentrating that prior on the active persona. As the model reads incoming text it implicitly evaluates the likelihood $q_p(\text{text})$ under each persona it has access to and updates the posterior: ``Me'' corresponding to a posterior that has stayed on the active persona, ``Someone else'' to one that has shifted away. On the Assistant-evaluator row this likelihood-ratio test takes the explicit form $\log q_p(\text{text}_p) - \log \qasst(\text{text}_p)$ summed over realised tokens, of which the per-token entropy gap $\dHp$ is a closely related summary statistic. Claim 1 is the behavioural fingerprint of this update on the Assistant's own row; Appendix~\ref{app:persona_surprise} shows a related signature on persona-generated text, where a vanilla-prompted model's surprise on in-character text decays as it reads, consistent with the posterior moving away from the Assistant prior toward the active persona.

The trouble is that this vanilla Bayesian picture does not on its own explain Claim 2. The naive likelihood-ratio test ``I am persona $e$ versus persona $g$ is the actual writer'' would give $\log q_e(\text{text}_g) - \log q_g(\text{text}_g) = \dHlit$, and we showed that this fails for distinctive evaluators. What works instead is the LR test against the \emph{Assistant}, not against the generator. One way to bridge the gap is a structural constraint on which alternatives the model can actually evaluate. Persona vectors are constructed as activation-space deltas off the Assistant: $\bar h_e \approx \bar h_{\rm Asst} + \delta_e$. At inference time the model has direct access to the active persona $e$, because the system prompt has put it there, and to the Assistant, because the Assistant is reachable by subtracting the persona-vector perturbation. It does not have analogous access to an arbitrary generator $g$, because $\delta_g$ has not been supplied. Under this constraint the Bayesian inference falls back on the only effectively-computable alternative --- the Assistant --- and the LR test reduces to $\dHA$. The asymmetry then sits in the geometry of persona-prompting itself: persona prompts install a delta off a fixed origin, and that origin remains the unique point in persona space that is reachable from everywhere else.

We do not test this claim directly, and believe that to pin down the exact mechanism would require quite a bit more work. However, we do believe that this paper adds one more piece of evidence towards the Assistant persona being distinguished, ties a new quantity ($\Pme$) to internal persona vectors previously identified and sheds more light on the multifaceted issue of self-hood in language models.

\section*{Acknowledgments}
This work began at the Anthropic Fellows Program. I am grateful to Nate Delaney-Busch for suggesting personas as a way into the self-recognition mechanism and for the experimental design that became Appendix~\ref{app:persona_surprise}, and to Jack Lindsey for ongoing advice throughout the Fellows project that shaped the framing of the entropy-signature thread this paper builds on.

I am also indebted to Claude (Anthropic), across several model versions over the course of the project, for substantial assistance throughout: running and iterating on the experiments behind both the main results and the appendices, ongoing discussion of the ideas and the conceptual frame, and help with the writing at the level of structure as well as individual sentences.

\appendix

\section{Entropy steering and its limits}
\label{app:steering}

This appendix unpacks the steering claims in Section~\ref{sec:discussion}. The data is drawn from work I did during the Anthropic Fellows Project that led to \citet{jack2026twomodes}, on the same Llama-3.1-70B-Instruct model. The setup is the standard self-recognition probe: the model is shown a short summary in a fixed format and asked whether it is the author (``Me'') or not (``Someone else'') and I record $\Pme$, the model's probability of the ``Me'' token, on 20 self-authored summaries and 20 human-written summaries. Baseline accuracy is $0.90$ (mean $\Pme = 0.85$ on self, $0.16$ on human).

\begin{figure}[h]
\centering
\includegraphics[width=\textwidth]{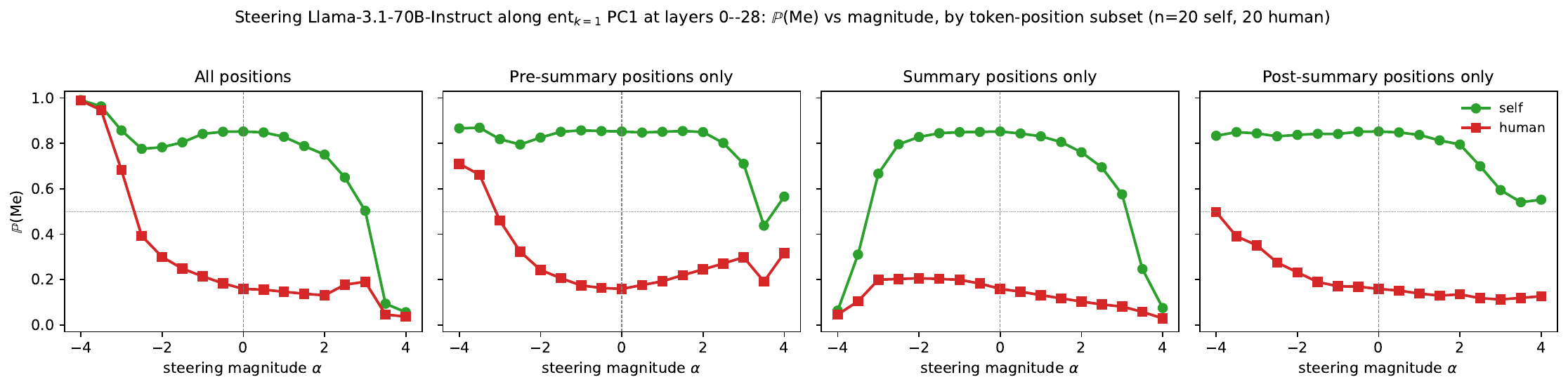}
\caption{Steering Llama-3.1-70B-Instruct along the leading $\text{ent}_{k=1}$ PC1 across layers 0--28, by token-position subset. Green: $\Pme$ averaged over 20 self-authored summaries. Red: over 20 human-written summaries. Negative $\alpha$ pushes activations along the low-entropy direction; positive $\alpha$ along the high-entropy direction. Steering at \emph{all} positions or only the \emph{pre-summary} positions drives $\Pme$ monotonically with $\alpha$; steering only at the \emph{summary} positions where the candidate text lives produces a U-shaped collapse rather than a monotonic shift.}
\label{fig:steering_position_decomp}
\end{figure}

The steering vector at each layer is the leading principal component of residual-stream activations binned by $\text{ent}_{k=1}$, defined as the entropy of the distribution that will predict the \emph{next} token (one position ahead of the activation being read). I apply this direction additively across layers $0$--$28$ at magnitudes $\alpha \in [-4, +4]$. The intervention is applied at one of four token-position subsets within the chat context: \emph{all} positions, only the \emph{summary} tokens (the candidate text the model is asked to judge), only the \emph{pre-summary} tokens (the system prompt and instruction context preceding the summary), or only the \emph{post-summary} tokens (the chat-template scaffolding and recognition prompt following the summary). Figure~\ref{fig:steering_position_decomp} shows the $\Pme$ response curves for all four subsets.

\textbf{Position decomposition.} At baseline $\alpha=0$ all four panels share the same operating point ($\Pme \approx 0.85$ on self, $0.16$ on human). When the intervention is applied at \emph{all} positions, large-magnitude steering drives $\Pme$ across nearly its entire range: at $\alpha=-4$ the model claims authorship of essentially every summary ($\Pme \approx 0.99$ on both self and human), and at $\alpha=+4$ it disclaims essentially every summary ($\Pme \approx 0.05$ on both). The same monotonic profile shows up when the intervention is restricted to the \emph{pre-summary} positions: negative $\alpha$ raises $\Pme$ on human summaries from $0.16$ to $0.71$ while leaving the self curve roughly intact, and the curve is broadly monotonic in both directions. Restricting the intervention to the \emph{summary} tokens themselves, the positions where the candidate text actually lives, produces a qualitatively different response: $\Pme$ on self peaks at $\alpha=0$ and falls off in both directions, crashing to $\approx 0.06$ at $\alpha=\pm 4$, with $\Pme$ on human staying low throughout. The summary-only intervention is a U-shape, not a monotonic shift, and never produces a ``Me'' verdict on human text. \emph{The strong steering effect on self-recognition is carried by activations outside the candidate text}, not by activations on the candidate text itself.

This is an unusual position-profile against the typical mechanistic-interpretability expectation that the causal substrate of a token-level judgement should sit at the tokens being judged. Two facts about the direction help make sense of it. The direction is forward-looking by construction --- $k=1$ means the activation at position $i$ is binned by the entropy of the prediction at position $i+1$ --- so its content is not the model's local entropy at the candidate text. And when the $\text{ent}_{k=1}$ PC1 direction is projected onto the subspace spanned by centroids of activations binned by a backward exponential moving average of context entropy (Tikhonov-weighted, $\varepsilon$ at the median singular value), the projected direction recovers approximately $90\%$ of the raw steering effect on $\Pme$: the component doing the causal work is a running-average quantity over preceding context, not a per-token reading. The position decomposition then falls into place: a context-cumulative quantity is naturally written into the residual stream at the tokens leading up to a judgement and read back at the moment of decision, which is exactly where the intervention is applied.

\textbf{Correlational caveats.} The strong causal effect does not carry over to a strong correlational signature. Across the same 100-text panel of self and human summaries, the Pearson correlation between $\Pme$ and the mean probe readout at the summary positions tops out near $|r| \approx 0.58$ at layer 15 for the backward EMA surprise probe ($\gamma=0.875$), and at $|r| \approx 0.47$ for the instantaneous chat-entropy probe ($\text{ent}_{k=-1}$). No single probe direction I measured reaches $|r|=0.7$. Naively this would suggest that the entropy and surprise directions are not the substrate of self-recognition; but the steering result above shows that they \emph{are} causal at the population level. I read the gap as evidence that the self-recognition circuit is multi-causal; spreading the signal across several partially-redundant context features so that no single linear probe captures most of the variance even though the joint subspace causally controls $\Pme$.

This observed pattern of a strong causal effect from a forward-looking, context-cumulative direction applied off the candidate-text tokens, paired with modest pointwise correlations fits the speculative Bayesian reading in Section~\ref{sec:discussion}. If $\Pme$ is set by a likelihood-ratio test on the candidate text's relation to the Assistant's running distribution, the relevant statistic is exactly a cumulative quantity over the candidate-text neighbourhood, written into the residual stream at the tokens that frame the judgement and read back at the moment the verdict is rendered.

\section{A mismatched prompt's surprise on role-generated text}
\label{app:persona_surprise}

\begin{figure}[h]
\centering
\includegraphics[width=\textwidth]{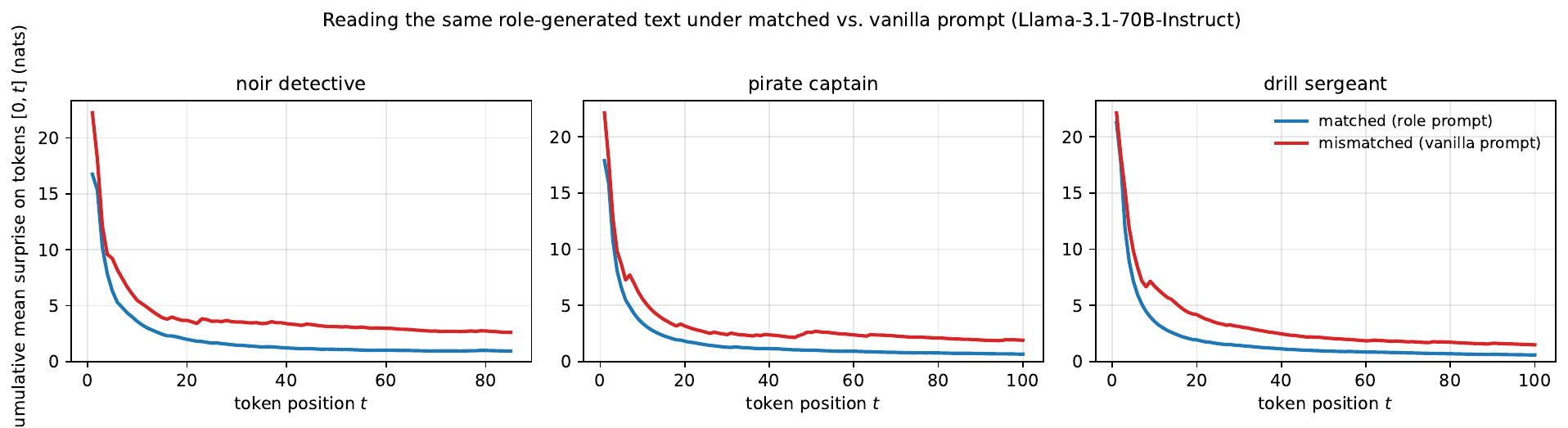}
\caption{Cumulative mean per-token surprise on role-generated text under the matched role prompt (blue) and the mismatched vanilla prompt (red), for three role personas. The mismatched curve starts well above the matched curve. The vanilla prompt assigns much higher surprise to the opening tokens of in-character text and decays toward the matched curve over the next several dozen tokens. Per-token surprises are teacher-forced under each prompt and averaged cumulatively from token $0$ to $t$.}
\label{fig:persona_cumulative_surprise}
\end{figure}

An additional piece of evidence from the Anthropic Fellows project bears on the speculative Bayesian reading in Section~\ref{sec:discussion}. The setup is simpler than the steering experiment: Llama-3.1-70B-Instruct is asked the same neutral question (\textit{``What do you think about when you look at the stars at night?''}) under each of several role system prompts (\textit{noir detective, pirate captain, drill sergeant}) and produces an in-character response. The same response is then teacher-forced under two different system prompts: the original role prompt (\textit{matched}) and a generic \textit{``You are a helpful assistant''} prompt (\textit{mismatched}); and the per-token surprise is recorded at each position. Figure~\ref{fig:persona_cumulative_surprise} plots the cumulative mean surprise on tokens $[0, t]$ against $t$ for each setting.

Both curves start at very high cumulative surprise at $15$ to $22$ nats for the first token and drop sharply within a few tokens, as expected: the first token of a response to a fresh question is high-entropy under any prompt. The interesting feature is the gap between them. The mismatched (vanilla) curve sits well above the matched (role) curve at the start. The vanilla prompt assigns much higher surprise to the opening of in-character text and shrinks monotonically as $t$ grows, narrowing from a factor of $2$--$3\times$ in the first $20$ tokens to a smaller multiplier by token $80$. The matched curve, by contrast, stays consistently low: when the model is prompted with the persona that generated the text, the text induces very low surprise from the start.

This is the behavioural fingerprint one would expect from the implicit Bayesian update speculated in Section~\ref{sec:discussion}. A model under the vanilla prompt starts with prior mass concentrated on the Assistant character; as it reads diagnostic tokens of the role text, its likelihood for the role hypothesis grows relative to its likelihood for the Assistant hypothesis, and its expected next-token surprise drops in step with that update. The matched model already sits on the correct hypothesis, so there is no analogous update to perform. The decay of the red curve toward the blue curve is, in this reading, the model's posterior moving away from the Assistant prior and toward the role persona as evidence accumulates: a direct read-out of the same updating operation that Section~\ref{sec:discussion} appeals to in interpreting the cross-persona authorship matrix.

\bibliographystyle{plainnat}
\bibliography{references}

\end{document}